\documentclass{ar-1col-S2O}
\usepackage[english]{babel}
\usepackage[numbers]{natbib}
\usepackage{amsmath}
\usepackage{url}
\usepackage{enumitem}
\usepackage{tabularx}

\usepackage{amsthm}
\newtheorem{definition}{Definition}
\usepackage{booktabs}
\usepackage{setspace}
\usepackage{graphicx}
\usepackage{makecell}

\jname{Submitted.}
\jvol{AA}
\jyear{2025}
\doi{10.XXXXX}

\usepackage[usenames,dvipsnames]{xcolor}
\usepackage{tcolorbox}
\usepackage{tikz}
\usetikzlibrary{plotmarks}
\usetikzlibrary{tikzmark}
\usetikzlibrary{calc}

\definecolor{myblue}{cmyk}{0.44, 0.2, 0, 0}
\definecolor{myyellow}{cmyk}{0, 0.13, 0.71, 0}
\definecolor{mypurple}{cmyk}{0.32, 0.47, 0, 0}

\newcommand{\bluesymbol}{%
        \begin{tikzpicture}[inner sep=0pt]%
        \node[shape=circle,fill=myblue,minimum size=2mm, inner sep=1pt] (char)
        {\ };
        \end{tikzpicture}%
    }

\newcommand{\yellowsymbol}{%
        \begin{tikzpicture}[inner sep=0pt]%
        \node[shape=circle,fill=myyellow,minimum size=2mm, inner sep=1pt] (char)
        {\ };
        \end{tikzpicture}%
    }

\newcommand{\purplesymbol}{%
        \begin{tikzpicture}[inner sep=0pt]%
        \node[shape=circle,fill=mypurple,minimum size=2mm, inner sep=1pt] (char)
        {\ };
        \end{tikzpicture}%
    }

\begin{document}

\markboth{Thompson, Candon, and V\'azquez}{The Social Context of Human-Robot Interactions}

\title{The Social Context of Human-Robot Interactions}

\author{Sydney Thompson,$^{*1}$ Kate Candon,$^{*2}$ and Marynel V\'azquez$^3$
\affil{$^*$Denotes equal contribution}
\affil{$^1$Computer Science Department, Yale University, New Haven, USA, 06511; email: sydney.thompson@yale.edu}
\affil{$^2$Computer Science Department, Yale University, New Haven, USA, 06511; email: kate.candon@yale.edu}
\affil{$^3$Computer Science Department, Yale University, New Haven, USA, 06511; email: marynel.vazquez@yale.edu}
}

\begin{abstract}
The Human-Robot Interaction (HRI) community often highlights the social context of an interaction as a key consideration when designing, implementing, and evaluating robot behavior. Unfortunately, researchers use the term ``social context'' in varied ways. This can lead to miscommunication, making it challenging to draw connections between related work on understanding and modeling the social contexts of human-robot interactions. To address this gap, we survey the HRI literature for existing definitions and uses of the term ``social context''. Then, we propose a conceptual model for describing the social context of a human-robot interaction. We apply this model to existing work, and we discuss a range of attributes of social contexts that can help researchers plan for interactions, develop behavior models for robots, and gain insights after interactions have taken place. We conclude with a discussion of open research questions in relation to understanding and modeling the social contexts of human-robot interactions.
\end{abstract} 


\maketitle

\section{INTRODUCTION}

The social context of human-robot interactions is key for the design, evaluation, and automatic generation of appropriate robot behavior \cite{bartneck2020human,thomaz2016computational}. It is generally accepted that the social context shapes how humans interpret signals \cite{vinciarelli2009social}, expect others (including robots) to act \cite{erel2024rosi}, and behave themselves \cite{aronson2016social}. In the field of Human-Robot Interaction (HRI), prior work typically investigates some element of robot's understanding of the social context (e.g., \cite{banisetty2021deep}), how robots should interact or  generate suitable behaviors within a particular social context (e.g., \cite{mutlu2006task,kim2025openvla}), and/or human experience with a robot acting within a social context (e.g., \cite{kim2014social}). Further, academic venues have long invited work that examines interactions in their social context~\cite{kiesler2004introduction}, calling for work that is ecologically valid \cite{hoffman2020primer} and addresses complex situated encounters~\cite{bohus2019situated}. 

Unfortunately, the use of the term ``social context'' in HRI is often overloaded or underspecified. For example, the term ``social context'' may refer to specific circumstances in which an interaction takes place or a particular application domain (such as healthcare or home robotics). To further complicate the matter, the term ``social context'' is often shortened to ``context'', which is even more overloaded in practice. ``Context'' can refer to information passed to an algorithm to interpret sensor observations (e.g., map of environment \cite{vazquez2022pose}) or to generate robot behavior (e.g., interaction states provide context to generate locomotion \cite{liu2016data}).
Papers in the HRI literature commonly use the term ``social context'' without describing what it means. Readers experienced in HRI can perhaps make an informed guess; but for others, like new HRI researchers, the term could get lost in translation, becoming jargon rather than clarifying the key ideas.
The lack of precision makes it difficult to identify, relate, and synthesize gaps in existing work about social contexts and their effects on human-robot interactions. What should we investigate next about the social context of a human-robot interaction?
What information about the social context should a robot use for reasoning when deployed in a new scenario? Without having a clear operational definition of what constitutes the social context of a human-robot interaction, it is unlikely that we will be able to answer these types of questions effectively.

Luckily, other related fields have grappled with the problem of conceptualizing the context of human-human and human-computer interactions. For instance, Social Psychology has long studied the concept of social context when investigating individual human behavior and human-human interactions. Stangor \cite{stangor2014principles} laid the foundation for the scientific study of how social factors influence individual and group behavior. Lewin \cite{lewin2013principles} posited that behavior is a function of individual characteristics and the surrounding social environment, $Behavior = f(Person, Environment)$, emphasizing the dynamic interplay between people and their social environments. More recent work by Argyle et al.~\cite{argyle1981social} has conceptualized social contexts for interactions in terms of social situations. Social situations are characterized by
the presence of two or more people who interact together and are shaped by a variety of factors, including the roles of the participants, the norms that govern behavior, people's goals or motives, and their forms of communication. As an example of the value of such conceptualization, social situations have recently been adapted to HRI, helping create more varied simulation environments for social robot navigation research \cite{tsoi2022sean}. 

In the field of Human-Computer Interaction, Dey \cite{dey2001understanding} provided an operational definition of context for ubiquitous computing: ``\textit{Context is any information that can be used to characterize the situation of an entity. An entity is a person, place, or object that is considered relevant to the interaction between a user and an application, including the user and applications themselves.}'' This definition served to scope context: For an indoor tour guide application on a mobile device, is the local weather context? According to the definition in \cite{dey2001understanding}, information about the weather is not context because the application is being used indoors and the weather does not affect it. However, information about the presence of other people who take part in the tour with the user is context, because they could affect which sites the user visits while using the application. Based on this notion of context, Dey went on to characterize context-aware computing, helping application builders more easily determine what features their applications should support and what context is critical to support the features. There is similar, practical value in better conceptualizing social context in the Human-Robot Interaction field. 

Riek and Robinson \cite{riek2011challenges} introduced an initial conceptual model of social context for researchers ``concerned with the automatic analysis of (and response to) human behavior''. They defined social context as: \textit{``the environment, E, where a person, P, is situated, with four factors that may influence P's behaviors. These factors include the situational context, P’s current social role in E, the cultural conventions of both E and P, and the social norms of E.''} O'Connor and Riek \cite{o2015detecting} expand upon this definition to provide formalisms for this social context and tease apart the conceptual ideas in practice, and Nigam and Riek apply this model to robotics \cite{nigam2015social}. Our proposed model builds upon these initial ideas, while focusing specifically on the social context of human-robot interactions.

In this survey paper, we first review existing definitions and the use of the term ``social context'' in HRI. The goal is to contrast different perspectives on how the community conceives of social context. Then, we propose an explicit definition of and conceptual model for the \textit{social context of a human-robot interaction}. Our conceptual model is inspired by the perspectives about context and social context described previously, but is designed specifically for HRI, considering the social nature of human-robot interactions, the importance of relationships within these interactions, and the unique characteristics of robots that differentiate them from other computing technology.
To explain our conceptual model, we  provide examples of how it can be applied to prior work in HRI and provide a taxonomy for socially contextual information that highlights the diversity of factors it can include.

Overall, with this review and our proposed conceptual model, we provide a pathway to think in an explicit and structured manner about  social contexts in HRI. We end by discussing practical uses of our conceptual model and open questions.
\section{REVIEW METHODOLOGY}
\label{sec:methodology}

We performed a systematic literature review with two main goals in mind: 1) better understanding the uses of the terms ``social context'' in the HRI literature (Sec. \ref{sec:socialcontext-in-hri}); and 2) defining and validating our proposed conceptual model for the social context of a human-robot interaction (Sec. \ref{sec:our-conceptual-model}).

We reviewed publications in well-established academic proceedings for HRI work from the years 2012-2023: ACM/IEEE Int'l Conf. on Human-Robot Interaction (HRI), IEEE Int'l Conf. on Robotics and Automation (ICRA), IEEE/RSJ Int'l Conf. on Intelligent Robots and Systems (IROS), IEEE Int'l Conf. on Robot and Human Interactive Communication (RO-MAN), and ACM Trans. on Human-Robot Interaction (THRI).

\begin{figure}[tb!]
    \centering
    \includegraphics[width=\linewidth]{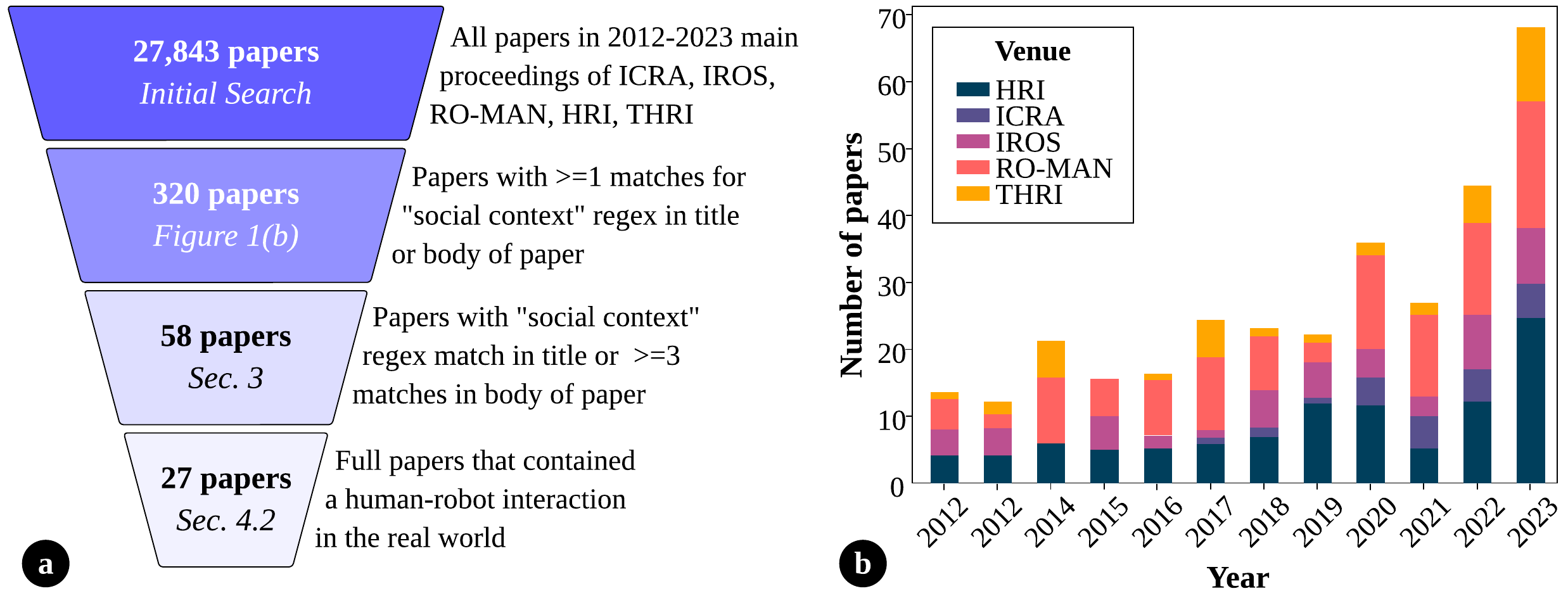}
    \caption{(a) Paper selection process for review. (b) Histogram of 320 papers using the term ``social context'' (or close variants like ``socially contextual'') in the title or body of the paper.
    Best viewed in color.}
    \label{fig:included_papers}
\end{figure}

The selected proceedings comprised of 27,843 articles, as shown in Figure~\ref{fig:included_papers}(a). We filtered the set of papers for work that discussed aspects of social context in human-robot interactions using string matching.
We searched for text matching the regular expression: \mbox{\texttt{social(?:ly)?[` '\-]?context(?:ual)?}}. The result was a corpus of 320 papers, a portion of which were short contribution papers because some conference proceedings combine short and full papers into the same volume. Figure \ref{fig:included_papers}(b) shows the number of papers in this corpus, organized by year. There is a clear increase in the use of the term ``social context'' and its close variants (per the regular expression), suggesting that the concept of social context is becoming increasingly important in HRI.

The 320 papers included many papers using the term ``social context'' without it being a particular focus of the paper; thus, we identified papers that had at least one match to the regular expression for ``social context'' in the title or had three or more matches in the body of the paper. This resulted in 58 papers, in which 257 sentences included the term ``social context'' in the papers' main content, excluding abstracts and titles.  In the next section, we analyze these 58 papers to understand how the robotics community uses the term ``social context''. Later, we narrow down this collection to papers that focus on real-world interactions to explain our proposed conceptual model for the social context.

\section{THE TERM ``SOCIAL CONTEXT'' IN HRI}
\label{sec:socialcontext-in-hri}

This section first discusses different explicit definitions that have been provided for the term ``social context'' among the 58 papers considered in our review that had at least one match
to the regular expression for “social context” in the title, or had three or more matches in
the body of the paper. Then, it discusses broader usage of the term in these 58 papers. 

\subsection{Explicit definitions of the term ``social context''.}
In the group of 58 papers, only 4 papers explicitly defined the ``social context'':

\begin{itemize}[leftmargin=*]
\item[--] Huang and Mutlu~\cite{huang2013repertoire} stated that the \textit{``social context or setting might characterize the physical environment (e.g., a domestic environment or a workplace), the organization of the interaction (e.g., dyadic interaction or group setting), the relative statuses of the participants (e.g., a supervisor or a subordinate), and the roles of the participants (e.g., a speaker or a bystander).''}

\item[--] Nigam and Riek~\cite{nigam2015social} defined \textit{``the social context for an agent (robot), \textit{P}, in a given environment, \textit{E}, as the disjoint union of several subsets: the situational context as a function of \textit{E}, the social role of \textit{P} in \textit{E}, \textit{P}'s cultural norms (irrespective of \textit{E}); \textit{E}'s cultural norms (irrespective of agents in \textit{P}); and the social norms for \textit{P} in \textit{E}.''}

\item[--] Zaech et al.~\cite{zaech2020action} stated that \textit{``social context... comprises of the positions and velocities of the other agents.''}

\item[--] Luber et al.~\cite{luber2012socially} used \textit{``the angle of approach $\alpha_{i,j}$ between the two [agents] $\pi_{i}$ and $\pi_{j}$ as the criterion to quantify and distinguish what [they] define here to be a social context.''}

\end{itemize}

Huang and Mutlu~\cite{huang2013repertoire} and Nigam and Riek~\cite{nigam2015social} suggested that the social dynamics between participants and the physical environment play a fundamental role in the social context of an interaction. For social robot navigation,  Zaech et al.~\cite{zaech2020action} and Luber et al.~\cite{luber2012socially} proposed that social context is comprised of the physical behavior of  agents, including their relative physical state. Together, the definitions from this subset of the HRI literature suggest:
\begin{enumerate}[leftmargin=*]
    \item Aspects of an \textit{environment} can be an important component of the social context of a human-robot interaction (e.g., home, work, cultural norms of a location, etc.).
    \item Attributes of individual \textit{agents} are an important part of the  social context (e.g., their conversational role, their position, their velocity, etc.).
    \item \textit{Associations} between agents can also be part of the social context (e.g., the relative orientation between two agents, job hierarchy, etc.).
\end{enumerate}

\begin{table}
\footnotesize
\caption{Categorization of the usage of the term ``social context'' in the HRI literature.}
\label{tab:sentence_stats}
\begin{tabularx}{\linewidth}{@{}>{\raggedright}p{1.42cm}p{1.5cm}Xp{2.4cm}@{}}
\toprule
\textbf{Category} & \textbf{Stats}\textsuperscript{\textdagger} & \textbf{Example} & \textbf{Citations} \\
\midrule
State of society & \makecell[tl]{4p  (6.9\%) \\7s  (2.7\%)} & {\footnotesize ``We believe that one key step HRI researchers can take to center the \textit{social context} is to include a societal implication consideration section in all papers'' \cite{ostrowski2022ethics}} & {\footnotesize \cite{ostrowski2022ethics,dobrosovestnova2023borrowing,winkle2023feminist,vsabanovic2014designing}} \\
\midrule
Domain & \makecell[tl]{3p  (5.2\%) \\3s  (1.2\%)} & {\footnotesize ``Robots were previously built to be used in \textit{social contexts} with members of the public, including healthcare, education, and robots used at home'' \cite{alhafnawi2022mosaix}} & {\footnotesize \cite{alhafnawi2022mosaix,chang2015interaction,pedersen2021using}} \\
\midrule
Task & \makecell[tl]{6p  (10.3\%) \\15s  (5.8\%)} & {\footnotesize ``\ldots{} goals are more important for a specific \textit{social context}.  For instance, if a robot were deployed in a service role that involved interacting with members of the public (e.g., museum tour guide, reception waiter, etc) \ldots{}'' \cite{briggs2017strategies}} & {\footnotesize \cite{10341925,coyne2023said,duffy2012suspension,van2023shake,bono2020social,briggs2017strategies}}\\
\midrule
Social Setting & \makecell[tl]{35p (60.3\%) \\104s (40.5\%)} & {\footnotesize ``The parallels between being excluded by robots and being excluded by humans \ldots{}, suggests that [robot-robot-human interaction] experiences have the potential to form a powerful \textit{social context} that impacts humans' emotions and behavior'' \cite{erel2021excluded}} & {\footnotesize \cite{nanavati2023unintended,nanavati2023design,tagne2016measurement,zaech2020action,10341925,luber2012socially,briggs2014actions,navarrosocial,van2023shake,tian2021taxonomy,chang2015interaction,lee2017steps,banisetty2021deep,arnold2017beyond,erel2021excluded,kim2022should,dobrosovestnova2023borrowing,erel2022carryover,liu2021avgcn,hwang2013developing,jin2019exp4,perez2018joint,chuang2020using,lee2019bayesian,tonkin2018design,oliveira2018friends,chita2019gender,grimm2023holistic,alhafnawi2022mosaix,arnold2018observing,moder2022proactive,bono2020social,xu2023solo,jackson2019toward,vazquez2017towards}}\\
\midrule
Physical Setting & \makecell[tl]{4p (6.9\%) \\4s (1.6\%)} & {\footnotesize ``\ldots{} our robot explored three types of social contexts on our college campus: study areas, dining areas, and lobby areas, across both the student center and library'' \cite{hayes2014avoiding}} &  {\footnotesize\cite{hayes2014avoiding,ostrowski2022ethics,tonkin2018design,xu2023solo}}\\
\midrule
Social \& Physical Setting (Explicit)& \makecell[tl]{19p (32.8\%) \\30s (11.7\%)} & {\footnotesize ``Both the limited roles of participants and the confines of the experimental environment present quite a different \textit{social context}  from that in which robots are eventually meant to operate.'' \cite{lee2017steps}} &  
{\footnotesize \cite{nanavati2023unintended,wang2019investigating,nigam2015social,coyne2023said,tian2021taxonomy,chang2015interaction,lee2017steps,banisetty2021deep,arnold2017beyond,erel2022carryover,liu2021avgcn,jin2019exp4,perez2018joint,papadakis2013social,kapoor2023socnavgym,joshi2017communal,vsabanovic2014designing,georgiou2023someone,chang2014observational}}\\
\midrule
Social \& Physical Setting (Implicit) & \makecell[tl]{33p (56.9\%) \\82s (31.9\%)} & {\footnotesize ``The researchers are immersed within the \textit{social context} they study, while being aware of the mutual influence researcher and participant have on each other and therefore keeping some distance to the people they study [22].'' \cite{nielsen2023using}} & {\footnotesize \cite{nanavati2023unintended,wang2019investigating,tagne2016measurement,foster2023social,campbell2020learning,coyne2023said,ostrowski2021long,briggs2014actions,duffy2012suspension,van2023shake,chang2015interaction,lee2017steps,banisetty2021deep,arnold2017beyond,kim2022should,dobrosovestnova2023borrowing,ostrowski2022ethics,ostrowski2022design,hennig2012expressive,kapoor2023socnavgym,pedersen2021using,nielsen2023using,winkle2023feminist,joshi2017communal,esterwood2021birds,tonkin2018design,oliveira2018friends,chita2019gender,georgiou2023someone,chang2014observational,xu2023solo,briggs2017strategies,jackson2019toward}}\\
\midrule
Other & \makecell[tl]{7p (12.1\%) \\12s (4.7\%)} & {\footnotesize  ``Discriminating and following others' gaze direction is an essential component of establishing a common \textit{social context} and it is pivotal to the ability to infer others' mental states'' \cite{perez2018joint}} \newline
{\footnotesize ``Support for this hypothesis would indicate that the agency ascribed to the robot is of key importance in determining whether human interactants will heed its protests, whereas lack of support would indicate the importance of other factors, such as \textit{social context}'' \cite{briggs2014actions}}& {\footnotesize \cite{briggs2014actions,duffy2012suspension,jin2019exp4,ostrowski2022design,perez2018joint,kapoor2023socnavgym,nielsen2023using}} \\
\bottomrule
\end{tabularx}
\vspace{1ex}
\begin{minipage}{\linewidth}
\footnotesize
\textsuperscript{\textdagger} The number and percentage of both papers (p) and sentences (s) using the term ``social context''. There are 257 sentences from 58 papers.\\
\end{minipage}

\end{table}

\subsection{Broader usage of the term ``social context''}
In the corpus of 58 papers, the term ``social context'' was used in a variety of ways without providing an explicit definition. Some usages of the term were broad, like using ``social context'' to describe a problem area or domain~\cite{chang2015interaction}. Other usages were  more specific, such as using ``social context'' as a synonym for ``social norm'' \cite{kim2022should}. Even within the same paper, authors sometimes used the term in different ways, referring to ``social context'' at different levels of specificity. Because usage varied so widely, it is difficult to suggest a single meaning or implicit definition underpinning the term in the current literature.

In order to better understand how people currently use the term ``social context'' in HRI, we systematically examined the uses of the term across the 58 papers. In an initial inspection, we found 6 common usages: \textit{state of society}, \textit{domain}, \textit{task}, \textit{social setting}, \textit{physical setting}, \textit{social and physical setting}. We then classified each of the 257 sentences that matched the regular expression for ``social context'' in the 58 papers into one of these 6 categories. Anything that did not clearly fit into one of these categories was classified as ``Other''.
Table~\ref{tab:sentence_stats} summarizes statistics and examples for each category, which we also describe below:
\begin{description}[wide, labelwidth=!, labelindent=0pt,itemsep=0.25em]
\item[State of society.] ``Social context'' refers to the broader state of society or of the world. 

\item[Domain.] ``Social context'' is used to explicitly refer to a broad HRI domain, like ``healthcare'' or ``entertainment''. The domain could include many different tasks or environments.

\item[Task.] ``Social context'' describes a specific application (e.g. robot tutoring), task (e.g. learning to read), or interaction scenario (e.g. robot asks a child to read something to it).

\item[Social setting.] The term ``social context'' describes the social setting, which might include beliefs, social norms, roles, expectations, or group membership.

\item[Physical setting.] The term ``social context'' describes the physical setting or environment of an interaction, without mentioning social aspects.

\item[Social \& physical setting.] ``Social context'' references both the social and physical environment of an interaction. In total, 41 papers used the term ``social context'' in this manner across 112  sentences. If both the physical and social setting were referenced explicitly, the paper was further categorized as ``social \& physical setting (explicit)'' (19 papers, 30 total sentences). If the references to the physical or social setting were implicit,  then the paper was  categorized as ``social \& physical setting (implicit)'' (33 papers, 82 total sentences). Implicit cases include situations where the physical environment was mentioned in nearby sentences, but not the sentence that contained the term ``social context''. 

\item[Other uses.] The intended meaning of the term ``social context'' was unclear or did not fit one of the above categories. 
\end{description}

The wide range of usages for the term ``social context’' motivated us to conceptualize a model, specific to human-robot interactions, that could serve to connect different perspectives in the literature.

\section{A CONCEPTUAL MODEL OF THE SOCIAL CONTEXT OF A HUMAN-ROBOT INTERACTION}
\label{sec:our-conceptual-model}

We propose a conceptual model for the \textit{social context of a human-robot interaction}. 
Our goal in creating this model was twofold. First, we wanted to provide an explicit definition of ``social context'' specifically for human-robot interactions that connects the literature.
We demonstrate how to apply our conceptual model to a variety of use-case scenarios and provide taxonomies for different types of attributes of social contexts drawing upon the literature. Second, we wanted to provide a tool -- the conceptual model -- to facilitate planning for interactions, generating behavior, and analyzing interactions after they have occurred. We discuss these practical implications in Section~\ref{sec:discussion} along with future research.

The proposed conceptual model is specifically designed and scoped to describe the social context of a human-robot interaction of interest, where the relevant human(s) and robot(s) act as embodied agents that perform actions in an environment, potentially influencing each other and changing the physical state of the world. For the purposes of the proposed model, a robot is embodied. Physical embodiment makes a robot inherently different from other computing technologies, as discussed in the book ``Human-Robot Interaction: An Introduction'' \cite{bartneck2020human}. While Reeves and Nass's Media Equation \cite{reeves1996media} suggests that people will act in a similar way with technology as with each other, this is not always true for robots~\cite{erel2024rosi}.

\subsection{Terminology}

Because our definition of the social context of a human-robot interaction is dependent on an interaction of interest, we first define what we mean by a human-robot interaction:

\begin{definition}
A \textit{\textbf{human-robot interaction}} is an exchange between agents, which must include at least one human and at least one robot. At its core, the interaction corresponds to a sequence of actions taken by the agents in a given environment, which are related to the task of the interaction (or the goals that each agent aims to accomplish). The interaction has temporal bounds that define when it begins and ends. 
\label{def:hri}
\end{definition}

Human-robot interactions can be \textit{explicit} (e.g., as in conversations, robot tutoring  settings, etc.) or \textit{implicit} \cite{ju2022design}, occurring without the explicit command or awareness of the human(s) involved in the interaction. For example, common implicit interactions in the social robot navigation literature involve having a robot navigate alone nearby people (e.g., as in \cite{mavrogiannis2019effects}). The people are not engaged in co-navigation or in an explicit information exchange with the robot, but they still adapt their actions to the robot as needed.

Human-robot interactions can be \textit{one-on-one} interactions between a robot and a human only, or they can be \textit{multi-party}, involving more agents  \cite{sebo2020robots,gillet2024interaction}. They can also change in size over time. The task or agent goals in an interaction can be the same for all agents or different. This can result in \textit{collaborative} interactions \cite{hoffman2004collaboration}, \textit{mixed-motive} interactions \cite{campos2016looking}, or \textit{adversarial} interactions \cite{duan2019robot}. Also, interactions can have varied length, from  brief interactions (like accidental encounters in a given physical space) to longer term (like a robot helping a person practice exercising). In general, we are not concerned with defining these aspects of interactions, but  let HRI practitioners and researchers decide what human-robot interaction is of interest, including the relevant agents, task, and temporal bounds. 

Given a human-robot interaction of interest, we define the social context of the interaction as follows:

\begin{definition}
    The \textit{\textbf{social context of a human-robot interaction}} is the set of attributes of the relevant agents, of their environment(s), and of their associations that influence the interaction. 
    \label{def:social-context-of-a-hri}
\end{definition}

Inspired by how Dey \cite{dey2001understanding} defines ``context'' for context-aware computing and existing uses of the term ``social context'' in HRI (Section~\ref{sec:socialcontext-in-hri}), we consider the \textit{attributes} in the social context of a human-robot interaction to be information that characterizes the relevant agents,  environment(s) and associations between them. The main requirement for these attributes to be part of the social context is that they influence the interaction either \textit{directly}, by having an effect on the sequence of actions of the interactants, or \textit{indirectly}, by having an effect in the interacting agents which, subsequently, influences their actions. 

The agents in Definition \ref{def:social-context-of-a-hri} can be of type \textit{robot}, \textit{person}, or \textit{other}. The robot(s) and human(s) that take part in the human-robot interaction of interest have attributes that are part of the social context of their interaction. For instance, this may include the embodiment of a robot, or a person's age, attention, etc. Other agents could include, for example, pets, whose attributes may be particularly relevant in home  or assistive robotics applications.

The inclusion of “associations” in Definition \ref{def:social-context-of-a-hri} is based on the importance that previous work places on relationships between agents when defining ``social context’’ for an agent in robotics~\cite{nigam2015social,o2015detecting}, as well as emphasis on these associations on context-aware computing~\cite{zimmermann2007operational}.
In general, associations between agents and environments in Definition \ref{def:social-context-of-a-hri} can be of three types: agent-agent, agent-environment, and environment-environment associations. Because there is important relational information that can affect human-robot interactions, we make these associations first-class entities in our conceptual model for the social context. That is, we give associations the same level of importance as agent and environment entities. Similar to the latter entities, associations can have  more than one attribute that is part of the social context of an interaction. For example, an association between a human and a robot could have information about roles (e.g., whether a person serves as a teacher for a robot) and human impressions about the robot (e.g., whether the person thinks the robot is acting competently). Also, an association between a person and an environment could contain information about the medium through which the person experiences that environment (e.g., in person or via a computer interface with a given set of attributes). 

\begin{figure}[tb!p]
    \centering
    \includegraphics[width=\linewidth]{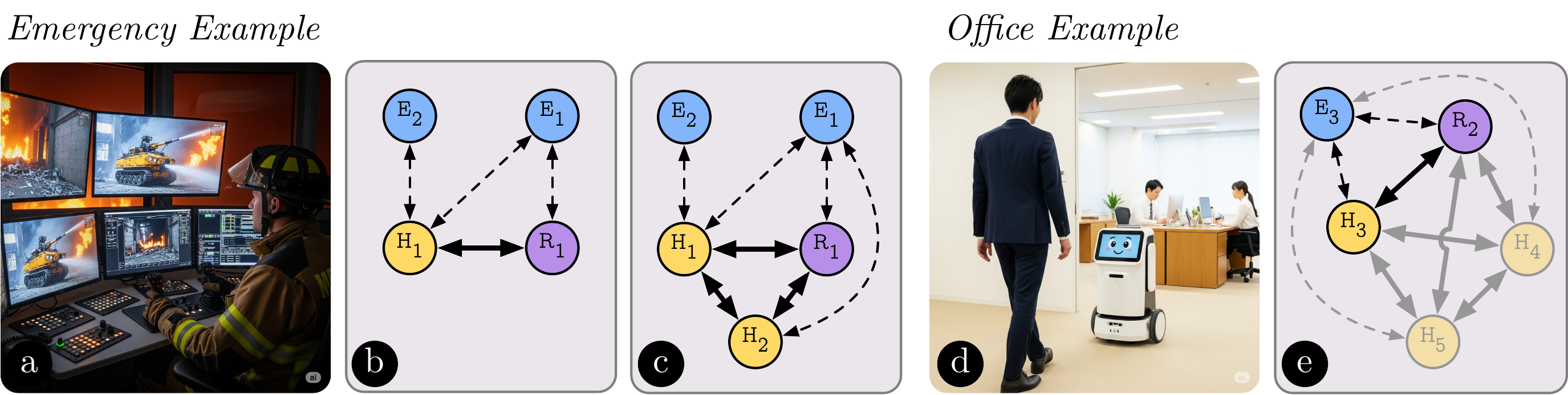}
    \caption{Examples of relevant entities for two different human-robot interactions: one between a firefighter and a robot, and another between a robot guide and a person in an office. Figures (a) and (d) illustrate the interactions at a given time; the images were created with Gemini's Generative AI. The diagram in (b) shows the relevant entities for (a), including the robot (\texttt{R$_1$}) and the firefighter (\texttt{H$_1$}). The firefighter teleoperates the robot that is in \texttt{E$_1$}, while being physically located in \texttt{E$_2$}.  Dashed edges have relevant agent-environment associations, while solid edges indicate relevant agent-agent associations. The diagram in (c) shows the interaction some time after (b), when the firefighter and robot team find another person (\texttt{H$_2$}). This illustrates how social context can be dynamic -- the relevant elements of the interaction can vary over time, so the attributes that influence human-robot interactions are also variable. In (d), a robot (\texttt{R$_2$}) brings a person (\texttt{H$_3$}) to an office, where there are two other people (\texttt{H$_4$} and \texttt{H$_5$}) who are not considered part of the interaction of interest. Hence, they are visualized in (e) with a lighter color-scheme. However, attributes of \texttt{H$_4$} and \texttt{H$_5$}, like whether they are attending to the \texttt{R$_2$} or \texttt{H$_3$}, influence the interaction of interest. As a result, the attributes of \texttt{H$_4$} and \texttt{H$_5$} are part of the social context of the interaction of interest. }
    \label{fig:firefighter}
\end{figure}

Across human-robot interactions, we find it  natural to think about the relevant entities of the interaction via graph visualizations. Nodes in the graph can be of two types (agents or environments) and edges can be used to encode associations between the nodes. If a node or an edge appears in the graph, it is because it has relevant attributes that are part of the social context. For example, imagine an emergency scenario in which a firefighter (\texttt{H$_1$}) teleoperates a robot (\texttt{R$_1$}) that goes into a disaster area with fire, as in Figure  \ref{fig:firefighter}a and Figure  \ref{fig:firefighter}b. The firefighter teleoperates the robot through a computer interface and from a remote location to stay safe. In this scenario, the social context of the interaction between \texttt{H$_1$} and \texttt{R$_1$} includes attributes from two environments. First, the environment of the robot (\texttt{E$_1$}), which the firefighter accesses via a teleoperation interface, influences the commands that \texttt{H$_1$} sends to the robot. Second, the environment in which the human is located (\texttt{E$_2$}) affects the human-robot interaction as well by influencing the human, e.g., to what extent it pays attention to the robot due to noise or other environmental distractions.

The attributes that are part of the social context of an interaction can change over time, and the relevant entities may also change. For instance, in the prior example of a firefighter-robot interaction, it could happen that the robot comes close to a victim (\texttt{H$_2$}), who then becomes part of the interaction as in Figure  \ref{fig:firefighter}c. The attributes that characterize this person and that influence the interaction are now part of the social context. 

There can also be additional agents that may not be part of the human-robot interaction of interest, but that have attributes that are part of the interaction's social context. For example, imagine that  the interaction of interest is between a robot \texttt{R$_2$} that guides a person \texttt{H$_3$} to a room in an office environment, where other people work (as in Figures \ref{fig:firefighter}d and \ref{fig:firefighter}e). Then,  whether these other nearby people attend to the interaction could influence the actions of \texttt{R$_2$} and \texttt{H$_3$}, e.g., making them speak in a lower volume. Thus, the attention of the other people is an attribute of the social context of the interaction of interest.

To more explicitly distinguish the possible attributes of all   environments and all agents from those that are part of the social context of a given human-robot interaction, we define:

\begin{definition}
    \textbf{\textit{Socially contextual information}} are the attributes in the social context of a human-robot interaction.
    \label{def:socially-contextual-information}
\end{definition}

In the reminder of this paper, we will often describe the social context of a human-robot interaction  as a collection, or set, of socially contextual information.

\subsection{Selection criteria for analysis of proposed definition of social context}
\label{sec:methodology-taxonomy}
To analyze the current literature through the lens of our proposed definition of social context, we created a subset of the papers selected for the literature review (per Section \ref{sec:methodology}) which focused on aspects of the social context of an interaction. In particular, for each paper in the corpus of 58 papers identified with our matching criteria for the regular expression for ``social context'', we read the abstract and checked the paper to determine if it was a full paper about a two-way interaction between at least one robot and at least one human. We did not include videos or extended abstracts in this subset of the papers because these shorter papers are typically about preliminary results. Also, we excluded  papers that only discussed online human-robot interaction studies, where participants did not experience an interaction with a robot in the real world. This filtering process led to 27 papers.

\subsection{Example use-case scenarios}
\label{ssec:use-case-scenarios}
This section illustrates how our proposed conceptual model for the social context of a human-robot interaction can be instantiated in specific scenarios. 
We categorized the 27 papers described in Section \ref{sec:methodology-taxonomy} based on how the paper's main contribution related to our concept of social context. The categories we identified were:\footnote{The numbers per category sum to 28 as one paper fell into both the Study and into the Computational Model and Systems groups.}
\begin{itemize}[leftmargin=*,itemsep=0.25em]
    \item[--] \textbf{Study:} 10 papers described a study in which socially contextual information constituted independent and/or dependent variables.
    \item[--] \textbf{Computational Models and Systems:} 9 papers described a computational model or system that used attributes of the social context of an interaction as inputs or estimated socially contextual information.
    \item[--] \textbf{Design:} 5 papers discussed the design of human-robot interactions. These discussions included both influences of the social-context of a human-robot interaction and how socially contextual information may be influenced.
    \item[--] \textbf{Survey:} 4 papers in our corpus were survey papers related to ideas captured by our definition of the social context of a human-robot interaction.
\end{itemize}

For the first three types of contributions described above, we discuss how our model fits the social context discussed in an illustrative paper. While not exhaustive, these use cases demonstrate the flexibility of our conceptual model and its applicability to existing work.

\subsubsection{Use case 1: Effects of robot appearance} 
\label{sssec:use-case-1}
In ``Actions Speak Louder Than Looks: Does Robot Appearance Affect Human Reactions to Robot Protest and Distress?'', Briggs et al. \cite{briggs2014actions} investigated whether a robot's appearance influenced how people responded to the robot verbally protesting a command. In this study, one robot built towers of colored cans. This robot was then removed, and participants were asked to instruct a second robot to knock down the can towers. 
The demolition robot protested the participant's request.

In applying our conceptual model to this study, we identified a variety of socially contextual information relevant to the interaction. First, the demolition robot's appearance was the independent variable in the paper, which they found to affect the human's perceptions about the robot's obligation to follow their commands. Thus, under our conceptual model, robot appearance is socially contextual information. Second, we consider the participant's commands to the demolition robot socially contextual information. Because the robot executes the participant's commands, the commands directly influence the interaction.
Third, the state of the can tower is socially contextual information attributed to the environment. This is because the cans being stacked into a tower impacts the participant's understanding of the task instructions to knock them down. Conversely, the color of the cans is likely not socially contextual information in this case (per Definition \ref{def:social-context-of-a-hri}) because there is no reason to believe that the cans' color affects the interaction.

Under our conceptual model, all positive results in HRI study papers, i.e., all confirmed factors that influence the interaction of interest directly or indirectly,  are socially contextual information. These factors are the independent variables of the study, e.g., characteristics of agents (such as action, role, appearance, physical state, or mental state), of environments (like location) or associations between agents and/or environments (like relationships). Negative results indicate that no significant evidence exists that an attribute is socially contextual information.

\subsubsection{Use case 2: Selecting listening behaviors} 
\label{sssec:use-case-2}
In ``A Bayesian Theory of Mind Approach to Nonverbal Communication'', Lee et al. \cite{lee2019bayesian} introduce a computational model for robot listening behaviors to indicate attentiveness. The robot's behavior is based on the storyteller's actions and a prediction of the storyteller's belief about the robot's attention.

A common assumption in computational models is that the inputs have an underlying causal relationship with the outputs. In this case, there are two models whose outputs are the storyteller's beliefs and the robot's action. For the models' inputs where the assumption holds, i.e. the causal relationship exists, the inputs are socially contextual information.\footnote{In practice, validating  causal effects for a particular  interaction can be difficult, and computational models can be negatively affected by spurious correlations in the data.}

Besides the behavior selection model, the robot's gaze and when it demonstrates a listening behavior is determined by a rule-based model that uses the storyteller's gaze, goals, and attributes of their speech including pitch, energy, pauses, and length. Because this rule-based model is encoding a causal relationship between the storyteller's behavior and the robot's actions, we consider these input attributes socially contextual information.

In addition to the models determining the robot's behavior, Lee et al. discuss specific attributes of the robot that, based on prior research, they think could influence the interaction, and carefully control them to prevent unwanted effects. Specifically, these attributes were robot's color, gaze, facial expressions, and utterances. These attributes are likely socially contextual information given prior HRI results.

\subsubsection{Use case 3: Designing interactions for an airport} In ``Design Methodology for the UX of HRI: A Field Study of a Commercial Social Robot at an Airport'', Tonkin et al. \cite{tonkin2018design} provide a methodology for designing  human-robot interactions in public environments that create a positive user experience. Tonkin et al. outline the steps for designing such an interaction and mention several factors about environments, humans, and robots that they believe could impact how an interaction would unfold. They note that for different deployment locations (e.g., airports, hospitals, or train stations) the volume level can influence whether users can hear any of the robot's speech. For humans, they mention that internal state and role (e.g., visitor or staff) can give important insight to their needs, which also may be dependent on the environment. For the robot they note that its morphology, personality, task, voice, identity, gestures, and screen display can impact how useful and positive the users' experience is. Under our conceptual model of social context, we would consider each of these attributes likely candidates for socially contextual information.

\subsection{Types of socially contextual information}
\label{ssec:hierarchical_contextual_info}
This section presents  taxonomies for socially contextual information. We built the taxonomies using papers described in Section \ref{sec:methodology-taxonomy}. For each of the 27 papers, we identified the key attributes of the human(s), robot(s), environment(s), and their associations that were explicitly discussed in the paper. Then, we used an iterative and collaborative process to develop the taxonomies using affinity diagrams. We expected the 27 papers to provide good coverage for the range of socially contextual information typically considered in HRI; but, when appropriate, we expand with other examples to better convey the richness of the social context of human-robot interactions.

\subsubsection{Environment attributes}
\begin{figure}[tb!p]
    \centering
    \includegraphics[width=\linewidth]{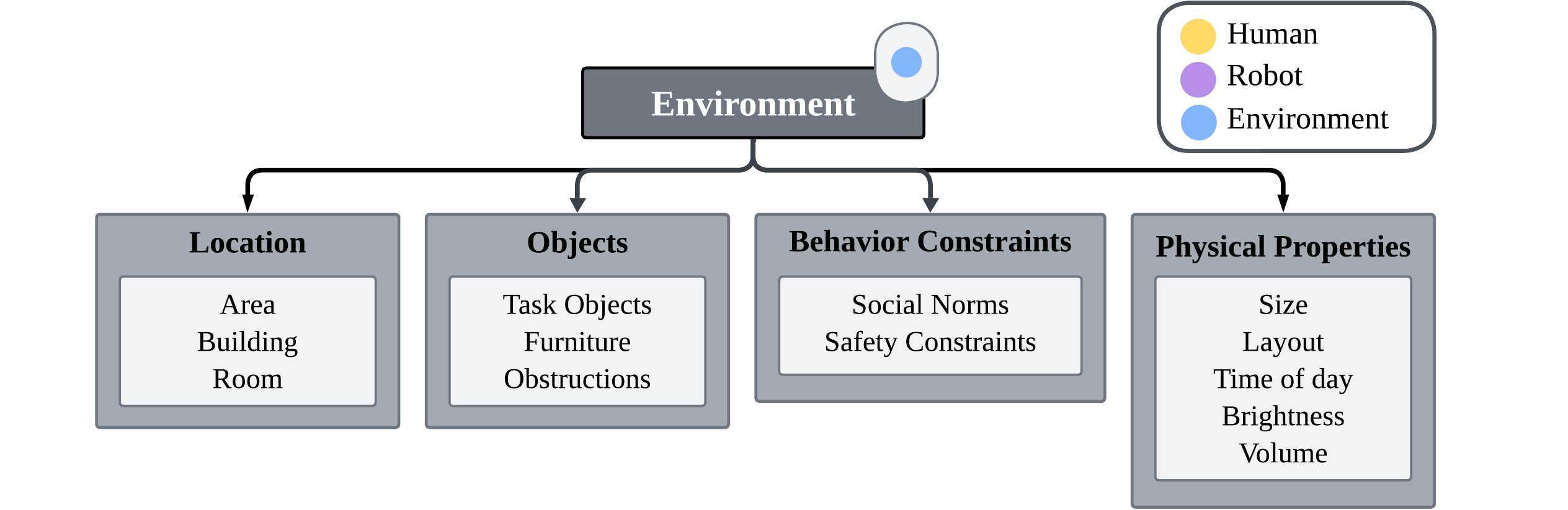}
    \caption[Taxonomy for socially contextual information of an environment.]{Taxonomy for socially contextual information of an environment. The blue symbol {\bluesymbol} is used to convey that these attributes could be relevant for the blue environment nodes in Figure~\ref{fig:firefighter}.}
    \label{fig:taxonomy_environment}
\end{figure}

As shown in Figure \ref{fig:taxonomy_environment}, we identified four main categories of socially contextual information describing the environment of a human-robot interaction: 

\begin{description}[wide, labelwidth=!, labelindent=0pt,itemsep=0.25em]
\item[Location.] The location was the most common type of socially contextual information for environments. Papers described locations at varying levels of granularity. For example, some  described the area in which interactions occur, such as public spaces \cite{tonkin2018design}. Other work focused on broader locations by noting the building in which an interaction takes place, such as a nursing home \cite{chang2015interaction} or grocery store \cite{nielsen2023using}. Even more specific were references to particular rooms, such as the lobby or activity area within an eldercare facility \cite{chang2014observational}.

\item[Objects.] Humans and robots often engage in physical interactions that involve the manipulation of objects, making attributes of such task objects an important piece of information that can influence the interactions. An example is the location of cans discussed in Section~\ref{sssec:use-case-1}. Similarly, humans and robots consider furniture  \cite{chang2014observational} or other obstructions \cite{jin2019exp4,chang2015interaction,xu2023solo,briggs2017strategies} when deciding how to navigate within an interaction.

\item[Behavior constraints.] Environments of human-robot interactions can have implications for the behavior of agents, which are socially contextual information that constrains their actions. Two noted examples from our literature review include safety constraints \cite{arnold2017beyond} -- which we view as hard constraints on the behavior of robots -- and location-specific social norms \cite{nielsen2023using} -- which we consider soft constraints. 

\item[Physical properties.] We consider the attributes of an environment that can be measured as their physical properties. When these attributes influence an interaction, they are socially contextual information. For example, papers referenced the layout of an environment as an attribute that can influence behavior. This layout was encoded via environmental maps \cite{xu2023solo} or referenced through elements such as hallway locations \cite{jin2019exp4}. Other physical properties include room size and condition (e.g., if it needs cleaning) \cite{lee2017steps}, the time of day \cite{nigam2015social},  the brightness of the environment \cite{tennent2017good}, or volume of an environment \cite{tonkin2018design}. Though not directly referenced in the papers in our literature review, this category of environment attributes could include other properties such as temperature or humidity.
\end{description}

\subsubsection{Agent attributes}
The papers that we reviewed discussed an agent's  actions,   attributes that are common population characteristics, physical state, appearance, and mental states. While some specific attributes were discussed more commonly for robots or for people in the selected papers, these five categories of socially contextual  information are generally applicable to both. Figure \ref{fig:taxonomy_agent} shows the derived taxonomy for agent attributes.

\begin{description}[wide, labelwidth=!, labelindent=0pt,itemsep=0.25em]
\item[Actions.]
Action-related attributes can  be categorized as either \textit{communicative}  or focused on \textit{strategy}. Communicative attributes can be associated to  \textit{verbal} and \textit{nonverbal} actions. The content of utterances \cite{van2023shake,perez2018joint,lee2019bayesian,oliveira2018friends} and tone of voice \cite{tian2021taxonomy,tonkin2018design, tennent2017good} comprised the most common verbal attributes. Lee et al. also note how a human's length of utterances or pauses in speech can influence the interaction \cite{lee2019bayesian}.
Attributes of nonverbal actions commonly include information about gaze \cite{chang2015interaction,vazquez2017towards,perez2018joint,lee2019bayesian}, facial expressions \cite{lee2019bayesian}, and gestures \cite{erel2021excluded,perez2018joint,hwang2013developing,arnold2017beyond,tonkin2018design,vsabanovic2014designing}.
Attributes related to an agent's strategy include an agent's goal \cite{banisetty2021deep,xu2023solo,briggs2017strategies}, their task \cite{arnold2017beyond,tonkin2018design,tian2021taxonomy,briggs2017strategies,nielsen2023using, campbell2020learning}, their planned trajectories \cite{banisetty2021deep}, the consistency of their behavior \cite{tian2021taxonomy}, or their recovery strategy \cite{tian2021taxonomy}. 

\item[Common Population Characteristics.] A variety of characteristics often used to describe populations -- also known as demographics -- were noted often for humans in the reviewed papers, although they could be applied to robots as well. For example, the papers that we reviewed noted the age \cite{chang2015interaction,lee2017steps,perez2018joint,nielsen2023using,pedersen2021using,lee2019bayesian,esterwood2021birds}, race \cite{chang2014observational}, gender \cite{lee2017steps,erel2022carryover,perez2018joint,chang2015interaction,pedersen2021using,lee2019bayesian,esterwood2021birds,vsabanovic2014designing,chang2014observational}, diagnoses (e.g., autism \cite{hwang2013developing}, dementia \cite{chang2014observational}, mental health conditions \cite{lee2017steps}), income \cite{lee2017steps}, education level \cite{lee2017steps}, and employment \cite{lee2017steps} of humans as factors that could influence interactions. While the selected papers did not discuss demographic characteristics of robots explicitly, it is common for people to assign gender and other similar attributes to robots as they anthropomorphize them. These attributes could influence robots' behavior (e.g.,  affecting how they communicate with people) or humans' behaviors  (e.g., affecting mental models of the robots).

\item[Physical state.] While details of the general location of a human-robot interaction can be captured in socially contextual information for the environment, the physical state of an agent is often a piece of socially contextual information. This state  often includes an agent's location, which can be represented in $(x,y)$-coordinates \cite{jin2019exp4}, but can have other abstractions. For instance, the physical state of a person could include their articulated body pose \cite{jin2019exp4,campbell2020learning} or orientation \cite{vazquez2017towards, lee2019bayesian}. Phyiscal state could also include motion information \cite{jin2019exp4}.

\begin{figure}[tb]
    \centering
    \includegraphics[width=\linewidth]{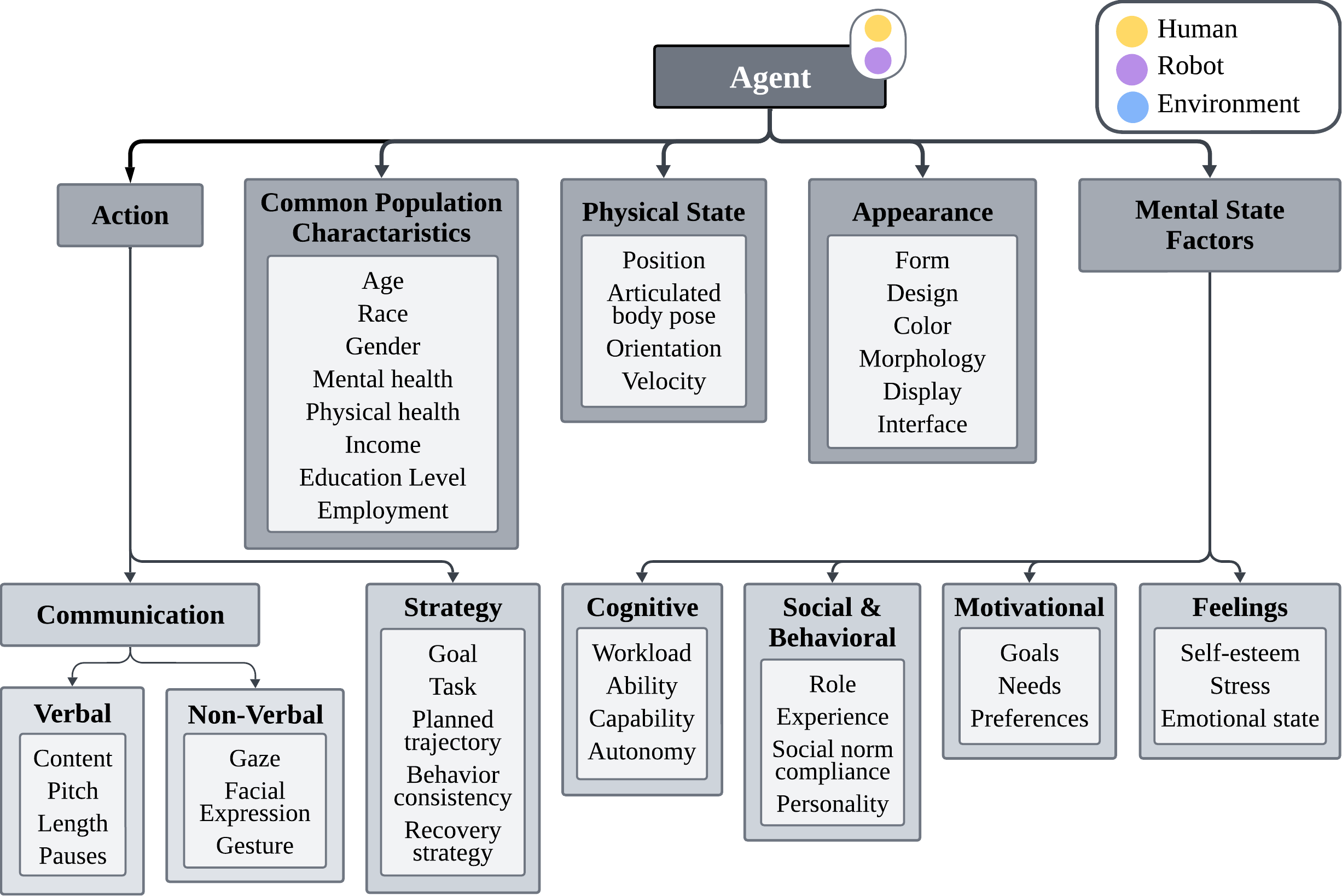}
    \caption[Taxonomy for socially contextual information of an agent]{Taxonomy for socially contextual information of an agent. The yellow  {\yellowsymbol} and  purple  {\purplesymbol} symbols  convey that these attributes could be relevant to the  human and  robot nodes in Figure~\ref{fig:firefighter}.}
    \label{fig:taxonomy_agent}
\end{figure}

\item[Appearance.] An agent's appearance can affect how other agents perceive them. The appearance of robots was often noted in the selected papers. For example, the general design \cite{chang2015interaction,pedersen2021using,tonkin2018design}, color \cite{lee2019bayesian}, and morphology \cite{tian2021taxonomy,lee2017steps,erel2021excluded,erel2022carryover,nielsen2023using,tonkin2018design,vsabanovic2014designing} of robots can affect interactions. Their screen display \cite{tonkin2018design, alhafnawi2022mosaix, nielsen2023using} or interface \cite{hwang2013developing} can also influence how people interact with them. While the papers that we reviewed did not explicitly discuss the appearance of humans, this attribute can influence human-robot interactions, e.g., in terms of whether a robot can (re-)identify a person visually. 

\item[Mental states.] Information about the internal, intellectual activity of agents is important socially contextual information because their internal states drive  behavior.\footnote{We considered mental states to be attributed to an agent when they were about the agent itself, not about other entities. Later, we discuss mental states about others (e.g., beliefs) as associations.} We identified four subgroups of attributes in the papers that we reviewed. First, mental states can be \textit{cognitive factors} such as cognitive workload \cite{tian2021taxonomy,vsabanovic2014designing}, abilities \cite{nielsen2023using, arnold2017beyond}, capabilities \cite{arnold2017beyond}, or level of autonomy \cite{lee2017steps}. Second, mental states can be \textit{social and behavioral} attributes. A common  factor is the role of an agent in an interaction. For example, whether a human is a resident, staff member, or visitor at an eldercare facility may influence how a robot chooses to interact with them \cite{tonkin2018design}. Similarly, whether a robot is an active or passive participant in a task (e.g., builder versus observer) can influence how a human might interact with said robot \cite{briggs2014actions}. Additionally, an agent's experiences (such as with technology \cite{lee2017steps, vsabanovic2014designing} or with robots in general \cite{chang2014observational}), an agent's compliance with social norms \cite{nielsen2023using}, or other personality traits \cite{esterwood2021birds, erel2022carryover} can also influence their actions in an interaction. Third, agents have \textit{motivational} states that can influence their actions, which include attributes related to goals \cite{briggs2017strategies,alhafnawi2022mosaix, xu2023solo, banisetty2021deep}, needs \cite{chang2015interaction, erel2021excluded}, and preferences \cite{vsabanovic2014designing}. Finally, mental states can be about an agent's \textit{feelings}, including  self-esteem \cite{erel2021excluded}, stress \cite{vsabanovic2014designing}, or other emotions \cite{nielsen2023using,tonkin2018design}.

\end{description}

\subsubsection{Association attributes} As shown in Figure \ref{fig:taxonomy_association}, attributes of associations generally fall into one of three categories:

\begin{figure}[t!p]
    \centering
    \includegraphics[width=\linewidth]{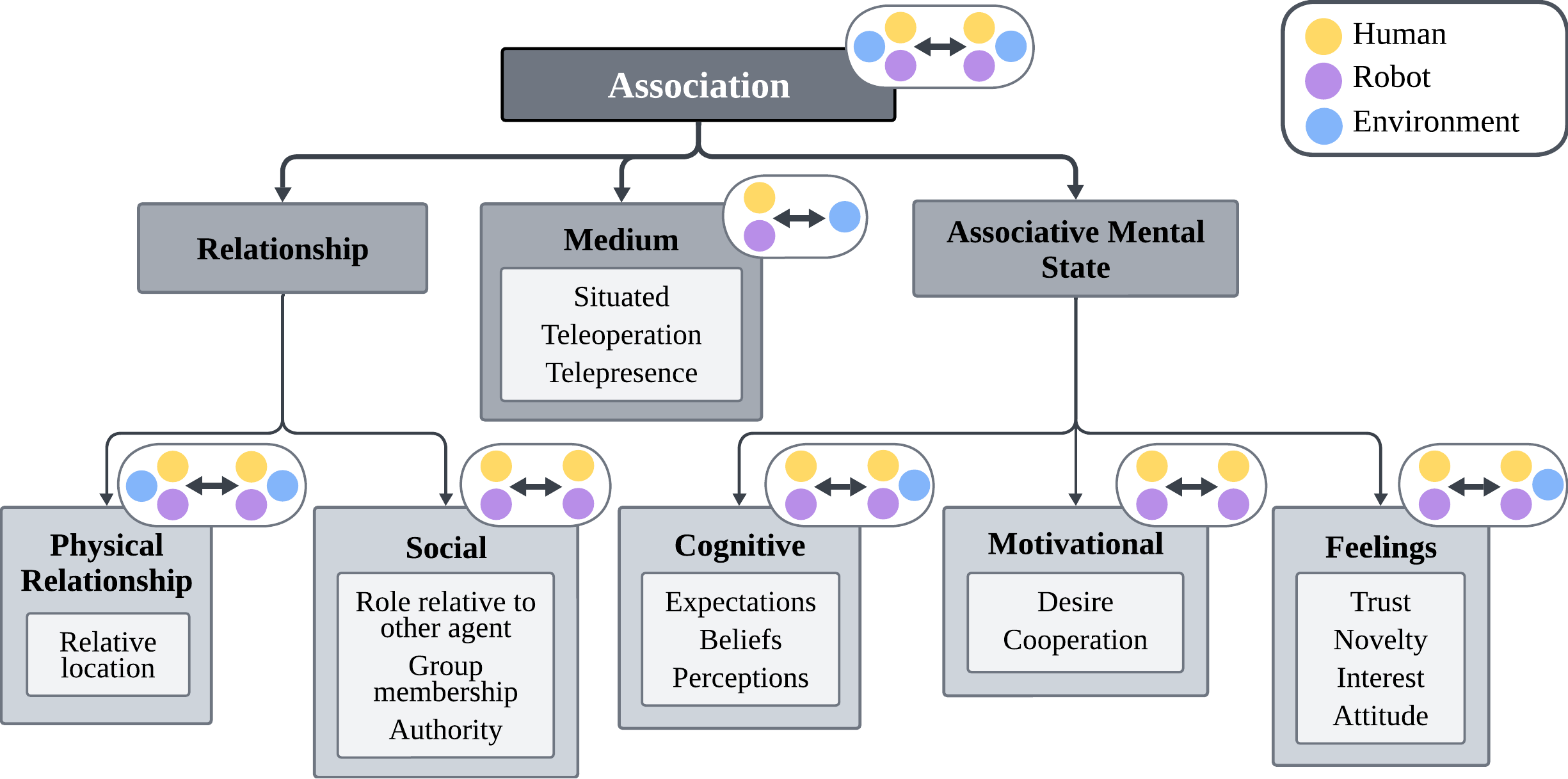}
    \caption[Taxonomy for socially contextual information of an association]{Taxonomy for socially contextual information of an association. Yellow {\yellowsymbol} and purple symbols {\purplesymbol} convey agent-to-agent associations. Blue symbols {\bluesymbol} convey associations that involve environments. These colors are consistent with Figure~\ref{fig:firefighter} and the prior taxonomies in Figure~\ref{fig:taxonomy_environment} and Figure~\ref{fig:taxonomy_agent}.}
    \label{fig:taxonomy_association}
\end{figure}

\begin{description}[wide, labelwidth=!, labelindent=0pt,itemsep=0.25em]
\item[Relationships.] Attributes about relationship associations can be of two types: they can encode information about \textit{physical relationships}  between any entities or information about \textit{social relationships} between agents. For example, Chang et al. described interactions in different rooms (e.g., lobby and activity room) in the same building, so the two environments had a physical relationship \cite{chang2014observational}. Papers also described how the distance between agents or between an agent and an object can influence the agent's actions \cite{xu2023solo,vazquez2017towards}. Attributes about social relationships include whether an agent has a human or robot partner \cite{oliveira2018friends}, the agent's group membership \cite{xu2023solo,arnold2017beyond}, or the authority agents have over one another \cite{arnold2017beyond}.

\item[Medium.] For agent-environment associations, the way in which the agent experiences the environment can influence their behavior and, hence, have an effect on human-robot interactions. For example, it is common for an agent to experience an environment by being situated in it, being co-located with other agents (as in Figure~\ref{fig:firefighter}d). Additionally, agents could experience an environment through computer interfaces,  such as a remote teleoperation interface for a robot (as in Figure~\ref{fig:firefighter}a), a video call, virtual reality,  etc.

\item[Associative mental states.] There are a variety of internal, mental states that involve more than one entity and can influence interactions. In the papers that we reviewed, we identified associative mental states that are \textit{cognitive} factors, such as expectations \cite{tian2021taxonomy,arnold2017beyond,nielsen2023using}, beliefs \cite{briggs2017strategies}, or perceptions \cite{briggs2014actions} about other agents or environments. Also, we identified \textit{motivational} mental states, e.g.,  whether a human  wants to interrupt a robot's task \cite{nielsen2023using} or touch a robot \cite{arnold2017beyond}, or thoughts about whether two agents are cooperating versus competing \cite{oliveira2018friends}. Associative mental states can also be about \textit{feelings}, such as trust in another agent \cite{tian2021taxonomy,arnold2017beyond}, novelty \cite{nielsen2023using}, interest \cite{nielsen2023using}, or attitudes \cite{erel2021excluded, erel2022carryover}. 
\end{description}

\section{DISCUSSION}
\label{sec:discussion}
Given the varied usage of the term ``social context'' in HRI, we proposed a conceptual model for the social context of human-robot interactions that can bridge  different perspectives. 
Now, we discuss ways in which we foresee  HRI practitioners and researchers leveraging this model in the future and highlight open challenges and interesting future research directions.

\subsection{Planning for human-robot interactions}

The taxonomies provided in Section~\ref{ssec:hierarchical_contextual_info} can serve as an initial checklist for thinking about different kinds of socially contextual information that could influence a human-robot interaction of interest.
First, outside of robotics, research has shown that interaction designers can have blind spots to novel conditions that come up during the deployment phase of a technology~\cite{patterson2009overcoming}, requiring contingency planning in the interaction design process. Likewise, this could  happen in HRI, where the situations that come up during a human-robot interaction can be novel and hard to predict \cite{murphy2008remote, honig2021expect}. 
Second, there can be unanticipated human behavior around the adoption of new technology. For instance, while a growing body of research suggests that robots can help support educational efforts
~\cite{belpaeme2018social}, research has also indicated that teachers can have negative attitudes towards education robots~\cite{reich2016robots}. Attitudes towards robots can be socially contextual information, as described by the taxonomy from  Figure~\ref{fig:taxonomy_association}, making it important to  plan ahead for them (e.g., by working with stakeholders to facilitate the introduction of robotics technology). Overall, by thinking about a variety of potentially-relevant socially contextual information ahead of an interaction based on the proposed conceptual model, HRI practitioners and researchers can prepare for novel situations and challenges that may come up in practice during human-robot interactions. 

A difficulty in planning for human-robot interactions is dealing with unknown socially contextual information. Prior work has explored understanding the social contexts of certain environments \cite{taylor2020situating} and developing systems that consider the social context during planning \cite{taylor2021social,suresh2023robot}. Our taxonomies of socially contextual information are not exhaustive as there can be more attributes that matter for a given interaction of interest than those reported in the papers that we reviewed. 
Further, some attributes in our taxonomies may not be relevant to all interactions, and people may change over time, inducing changes in the factors that influence their interactions with robots. For these reasons, we advocate for iterative interaction design processes (e.g., as in \cite{lee2009snackbot,hoffman2015design,matheus2025ommie}), where interactions are repeatedly prototyped, tested, and refined. Additionally, because user testing can be slow and expensive, it is important for the HRI community to continue innovating in design and evaluation methodologies which can accelerate understanding of social contexts. These methodologies may include the use of simulations \cite{hoffman2014designing,tsoi2022sean}, virtual and augmented reality technologies \cite{suzuki2022augmented,walker2023virtual},  front-end human-robot interfaces \cite{toris2015robot}, 
crowdsourcing \cite{lee2021interactive,tsoi2021approach}, etc. While these methodologies may not fully replicate real-world results \cite{li2019comparing,tsoi2024influence,esterwood2025virtually}, they can accelerate the identification of socially contextual information and, hence, help develop better interaction paradigms and  robust robotics technology.

\subsection{Robot behavior generation during interactions}
The proposed conceptual model for the social context of human-robot interactions can aid in developing autonomous robot behavior. Previous work has shown that adapting a robot's behavior to different contexts can improve user experience \cite{porfirio2020transforming,mutlu2006task}. Intuitively, imagine that a robot had a computational model of the social context of a human-robot interaction -- that it understood what attributes of the relevant agents, environment(s) and of their associations influenced the actions of the agents of interest. If the robot could predict the outcome of these influencing effects, it could then generate suitable behavior by searching over its action space for the best actions that help it achieve a desired outcome. This behavior generation setup is generally intractable but approximations have found value in HRI, e.g., via receding horizon planning or optimization ~\cite{ng2023takes,ghose2024planning,renz2024moving}.

Computationally representing social contexts and, further, learning the dynamics of interactions in a way that captures all relevant socially contextual information is 
 a difficult challenge. 
First, a particular interaction may contain extensive amounts of socially contextual information. Foundational work on computationally modeling the social context of an agent in robotics utilized symbolic representations~\cite{nigam2015social, o2015detecting}. While effective at describing varied contexts, such representations can be potentially prohibitive from a space requirement due to the explicit nature of the symbolic abstractions. Given recent advances in representation learning with neural networks, there is opportunity for implicit representations of social contexts to be more scalable. But HRI data is scarce, making the use of inductive learning biases likely necessary for effective generalization~\cite{mitchell-theneedforbias}. In particular, we hypothesize that  utilizing relational abstractions or graphs (as in Figure~\ref{fig:firefighter}) and machine learning models designed specifically to reason about these abstractions, like Graph Neural Networks~\cite{battaglia2018relational}, could be beneficial for HRI given the importance of associations between agents and environments in the social context of an interaction. Indeed, recent work  in HRI has utilized graphs as abstractions to encode  attributes of the environment and team dynamics~\cite{altundas2022learning},  for learning cost functions for motion planning~\cite{manso2021graph}, and as state abstractions for learned social robot behavior policies~\cite{gillet2025templates}. 

Second, it is not clear what the best level of specificity is for computationally abstracting socially contextual information. For instance, consider a person's utterance. It could be computationally abstracted as a high-level intent, text, or a sound wave. Which abstraction is more useful in practice depends on what the robot is trying to achieve during an interaction. For example, high-level intent could be useful for coordinating the robot's behavior with the user~\cite{huggins2021practical}. The sound wave could aid in synchronizing the robot's speech with the user to build rapport
~\cite{nishimura2021vocal}.  Thus, we suspect that  general and efficient computational abstractions for social contexts will ultimately need to be hierarchical. In alignment with this hypothesis, the spatial reasoning community in robotics has advocated for hierarchical, metric-semantic environment maps to enable complex physical robot behavior~\cite{hughes2024foundations}. 

Third, important  socially contextual information is not directly observable by robots, such as internal mental states. While machine learning techniques are fueling a variety of approaches for inferring internal human states, e.g., from affective states~\cite{aly2012towards,staffa2022enhancing,churamani2022affect} to perceptions of robot behavior~\cite{cui2021empathic,stiber2024uh,zhang2025predicting}, it remains difficult to measure these internal states in a scalable manner. This poses challenges for building datasets on which to train models and evaluating prediction performance in practice.

Given these challenges and recent advancements in generative Artificial Intelligence, it may seem natural to resort to large neural network models, like Vision-Language Models (VLMs) to generate socially-contextual robot behavior. For robot manipulation and navigation, VLMs built on large-scale Internet data are serving as effective backbones for improved scene understanding, and bridging the gap between high-level instructions and low-level control (e.g., see~\cite{vuong2023open,song2024vlm}). In HRI, large models are becoming increasingly popular to create models of humans~\cite{zhang2023large}, generate more varied robot speech~\cite{skantze2025applying,kim2024understanding}, and implement a variety of functionality in the control system of a robot~\cite{williams2024scarecrows}. However, there is limited data capturing subjective human feedback that can be used in HRI to learn end-to-end robot policies with large models. Additionally, though large models seem to be ever-improving, their reasoning remains opaque, making it difficult to understand why, or even when, these models make mistakes. This motivates incremental learning approaches focused on continued improvement of robot autonomy~\cite{churamani2020continual} as well as utilizing a variety of human feedback~\cite{chetouani2021interactive,chernova2022robot,fitzgerald2022inquire} to adapt or steer robot behavior policies.

\subsection{Post-interaction analyses}

The proposed conceptual model for the social context of human-robot interactions can help to understand interactions after they have taken place. For example, one could analyze the appropriateness of robot behavior based on the relevant social norms that apply to them, which are part of the social context. This idea is in line with prior work in social robot navigation, which has categorized social situations to identify types of interactions where robot performance needs improvement~\cite{tsoi2022sean}, investigated how organizational factors affect the way people respond to delivery robots~\cite{10.1145/1349822.1349860}, or classified the environment to adapt robot behavior to different social norms~\cite{banisetty2021deep}. Furthermore, our taxonomies for socially contextual information can help HRI researchers think about potential confounds that could lead to incorrect conclusions in experimental HRI work because, by definition, the attributes of the social context influence interactions. Finally, our conceptual model could also aid in defining the concept of ``generalization'' in HRI. Research in Human-Robot Interaction has called for building a generalizable theory of HRI from ``in the wild" social encounters~\cite{jung2018robots}: \textit{``a principled understanding of what to expect with different types of robots, performing different types of tasks, in different types of social situations and cultures."} Our conceptual model for the social context of human-robot interactions can serve to establish a broadly-applicable notion of generalization for such theories, where effective generalization entails predicting accurate outcomes in novel social contexts. These novel  contexts are characterized by novel values  for known socially contextual information, as well as by completely novel environment, agent, and association attributes that influence an interaction of interest. 

It would be transformative if robots could reason about causal effects in social encounters based on their observations of interactions and new data that they collect. Recent work demonstrates the feasibility of learning causal relationships within the HRI domain given a known set of relevant features \cite{edstrom2023robot, castri2022causal, castri2023enhancing}, but more work is needed in this direction to capture a wider range of socially contextual information.
Ultimately, causal competency, including understanding the social context of human-robot interactions, may be necessary for autonomous, social robots to behave ethically in novel situations \cite{hellstrom2021relevance}.

\section*{DISCLOSURE STATEMENT}
The authors are not aware of any affiliations, memberships, funding, or financial holdings that might be perceived as affecting the objectivity of this review. 

\section*{ACKNOWLEDGMENTS}
We are thankful to Nathan Tsoi for helping us create the collection of  papers considered in this review. We are also grateful to Anind Dey, Laurel Riek, Bilge Mutlu, Sasha Lew, Houston Claure, and Ulas Berk Karli for their feedback on preliminary drafts. This work was partially supported by the  National Science Foundation (NSF) under Grant No. (IIS-2143109) and (IIS-2106690), and the U.S. Air Force Offce of Scientifc Research (AFOSR) under the Young Investigator Program (Award No. FA9550-24-1-0085).

\bibliography{refs}

\end{document}